\documentclass[conference]{IEEEtran}
\IEEEoverridecommandlockouts
\usepackage{algorithm,algorithmic}
\usepackage{cite}
\usepackage{fancyhdr}
\usepackage{amsmath,amssymb,amsfonts}
\usepackage{algorithmic}
\usepackage{graphicx}
\usepackage{textcomp}
\usepackage{xcolor}
\usepackage{hyperref}
\usepackage{float}
\usepackage{fancyhdr}
\usepackage{lastpage}
\usepackage{eso-pic}
\usepackage{fancyhdr}
\usepackage{tikz}
\usepackage{amsmath}
\def\BibTeX{{\rm B\kern-.05em{\sc i\kern-.025em b}\kern-.08em
    T\kern-.1667em\lower.7ex\hbox{E}\kern-.125emX}}
\begin{document}

\title{DTization: A New Method for Supervised Feature Scaling}

\author{Niful Islam\\
Department of Computer Science and Engineering, United International University\\ United City, Madani Avenue, Badda, Dhaka 1212, Bangladesh\\ Email: nislam201057@bscse.uiu.ac.bd}

\maketitle

\begin{abstract}
Artificial intelligence is currently a dominant force in shaping various aspects of the world. Machine learning is a sub-field in artificial intelligence. Feature scaling is one of the data pre-processing techniques that improves the performance of machine learning algorithms. The traditional feature scaling techniques are unsupervised where they do not have influence of the dependent variable in the scaling process. In this paper, we have presented a novel feature scaling technique named DTization that employs decision tree and robust scaler for supervised feature scaling. The proposed method utilizes decision tree to measure the feature importance and based on the importance, different features get scaled differently with the robust scaler algorithm. The proposed method has been extensively evaluated on ten classification and regression datasets on various evaluation matrices and the results show a noteworthy performance improvement compared to the traditional feature scaling methods.
\end{abstract}


\section{Introduction}
Machine learning (ML), a sub-field of artificial intelligence, focuses on analyzing the data, finding patterns, and making predictions without explicit programming. ML encompasses a wide range of algorithms that are broadly classified into four categories that are supervised, semi-supervised, unsupervised, and reinforcement learning \cite{mirzaei2022machine}. The supervised machine learning algorithms learn from the labeled training data for making predictions on the unseen data. The training data consists of input features and corresponding output labels. The goal of the supervised machine learning algorithm is to construct a function that can generalize the training data and make necessary predictions \cite{naude2023machine}. Supervised learning has been successfully incorporated in various domains such as decease detection \cite{islam2023toward}, weather forecasting \cite{markovics2022comparison}, and many more. Unsupervised learning also known as clustering, on the other hand, works with the unlabeled data \cite{wen2022survey}. Unlike supervised learning, clustering does not have any predefined output label. Instead, the algorithms look for underlying structures, relationships, or patterns in the data. The common applications of unsupervised learning can be seen in customer segmentation \cite{alghamdi2023hybrid}, recommendation systems \cite{roy2022systematic}, anomaly detection \cite{chatterjee2022iot}, and many more. \par 
Decision tree (DT) is a popular supervised machine learning algorithm that incorporates a hierarchical structure to make predictions based on the input data \cite{islam2023knntree}. The algorithm constructs a tree data structure where the internal nodes represent a feature and leaf nodes constitute class labels. Decision trees are built by recursively partitioning data based on the most informative features, with the goal of maximizing information gain or minimizing impurity at each step. Therefore, the important features lie on top of the tree whereas the less important features are on the bottom of the tree. \par
In machine learning, data pre-processing is generally applied before training the machine learning model to enhance the effectiveness of the model. Feature scaling is one of the popular and most useful data pre-processing techniques for both supervised and unsupervised learning that aims to ensures that all features are on a similar scale. Feature scaling is particularly useful for those machine learning algorithms that use distance measurement for constructing the model such as K-nearest neighbors, linear regression, k-means clustering \cite{saheed2022machine}. There are several types of feature scaling methods available. The most common feature scaling technique is min-max scaling or normalization which scales the features into a specific range that is usually between 0 to 1 \cite{mazziotta2022normalization}. Another common feature scaling method is Standardization, also referred as Z-score normalization which transforms the data into a standard Gaussian distribution. Log transformation is another highly effective feature scaling method that applies logarithmic function on the input features for reducing the skewness of the data \cite{singh2022lt}. Lastly, robust scaling is one more method that is robust to the outliers. \par 
In this paper, we present a novel feature scaling technique named DTization that employs decision tree and robust scaler to construct a powerful feature scaling algorithm. Since DT arranges features based on the correlation with the output variable, this algorithm can be employed to find out the most useful features. Therefore, we rank the features based on the priority obtained from DT and map the priority with a scaling factor. While scaling the features, we multiply the scaling factor with the output obtained from the robust scaler. We have evaluated the proposed method on classification and regression datasets and the results show an outstanding improvement in the performance compared to other feature scaling techniques. In summary, the research article the following contributions:
\begin{itemize}
    \item A novel method for supervised feature scaling.
    \item The algorithm has been evaluated on ten classification and regression datasets on various evaluation matrices and the results show a noteworthy performance improvement.
    \item A comparison with four feature scaling techniques for justifying the effectiveness of the algorithm.
    
\end{itemize}
The remaining sections of this article are structured as follows. Section \ref{sec:rw} presents a summary of the relevant literature. Section \ref{sec:method} presents the proposed algorithm of feature scaling. Subsequently, Section \ref{sec:results} discusses the results obtained from the algorithm and makes a comparative analysis with other feature scaling techniques. Lastly, the article concludes in \ref{sec:conclusion}.

\section{Related Work}
\label{sec:rw}

There is a limited number of recent articles proposing a new feature scaling technique. However, researchers have experimented with different feature engineering methods for increasing the generalization ability of the training data. Ahsan et al.\cite{ahsan2021effect} presented a comparison between eleven machine learning algorithms and six data scaling methods for finding the best pair for heart disease classification. According to the experiment, classification and regression trees (one form of decision tree) produce the best performance when the dataset is scaled with robust scaler and quantile transformer. Zhang and Zhao \cite{zhang2017fetal} experimented with cardiotocographic data for assessing fetal state. They combined principle component analysis (PCA) for feature selection with AdaBoost algorithm to achieve high performance. Devi et al. \cite{devi2021variant} also conducted an experiment for detecting hypothyroid disease. They selected a dataset with twenty-four features and employed two dimensionality reduction techniques named principal component analysis and linear discriminant analysis pairing with various machine learning techniques for hypothyroid disease classification. According to the experiment, support vector machine (SVM) performs the best with principle component analysis and logistic regression, Naive Bayes pairs the best with linear discriminant analysis. Md et al. \cite{md2023enhanced} investigated four feature scaling methods and three feature selection methods for liver disease detection. The feature scaling methods explored in the article are standard scaling, max scaling, min-max scaling, and robust scaling. With a total of twelve preprocessed datasets, extra tree classifier performed the best for disease detection. Tripathi et al. \cite{tripathi2021feature} performed an investigation to find the best methods for malware detection systems. According to the result, robust scaler for feature scaling along with chi-square for feature selection produces the best result on random forest classifier. Shekhar et al. \cite{shekhar2022pknn} employed mutual information for finding the best features in a dataset and the selected features are then used to train K-Nearest Neighbor classifier. Mutual information finds the correlation between two variables and outputs a value in a range of zero to one. Albulayhi et al. \cite{albulayhi2022iot} presented a novel feature selection technique for intrusion detection. This technique leverages information gain and gain ratio to select some number of top features. On the selected features, they apply mathematical set theory to determine the final features that are used for final model training. The proposed model was found to largely impact the classification of the model on two intrusion detection datasets.\par 
To summarize, there have been many researches for finding the best feature engineering techniques for various machine learning problems. However, there has been very little research about developing a new algorithm for feature scaling.

\section{Methodology}
\label{sec:method}
\begin{figure}
    \centering
    \includegraphics[width=9cm,height=7cm]{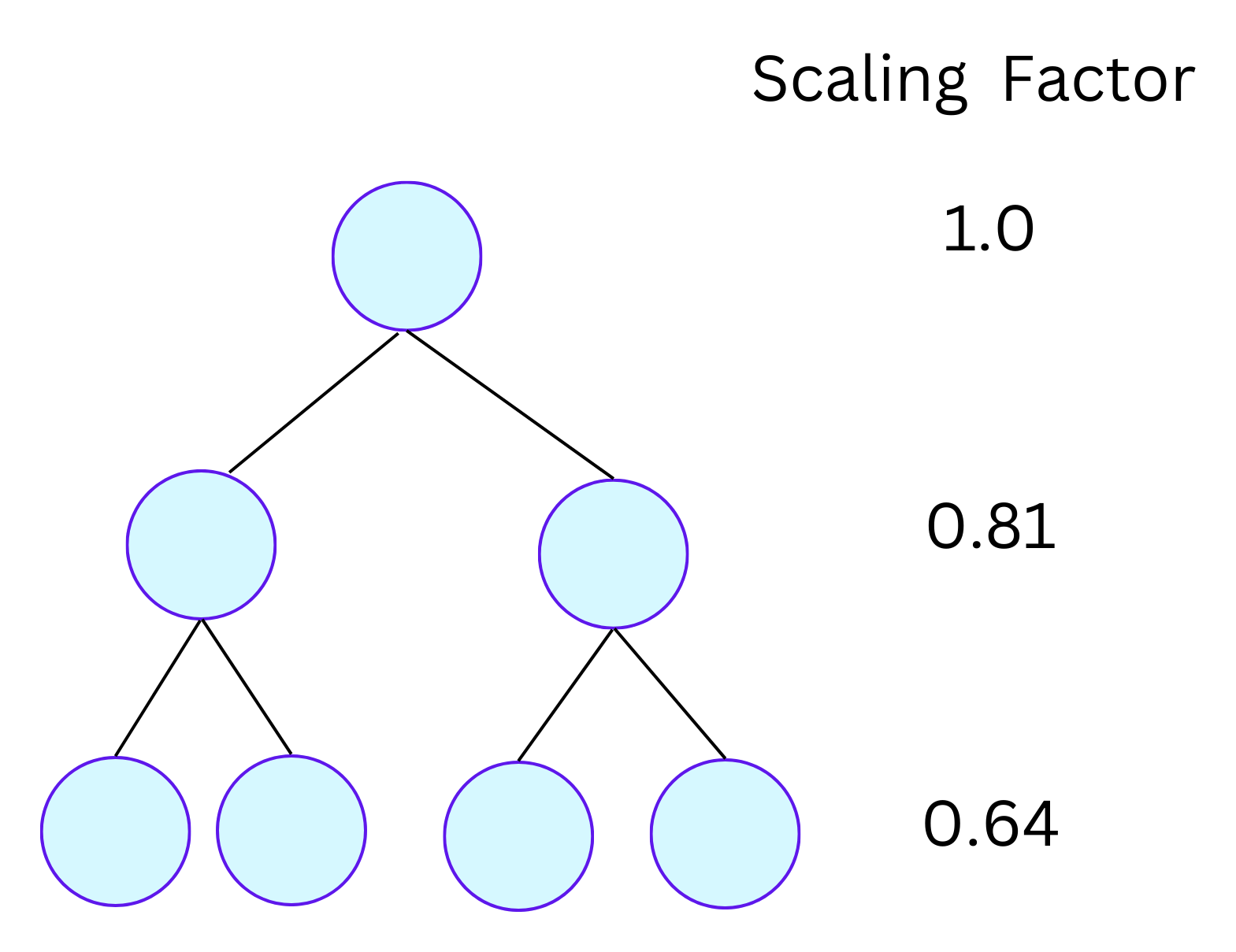}
    \caption{Calculating scaling factor for each features}
    \label{fig:scaling-factor}
\end{figure}

The proposed algorithm can be divided into two main parts that are calculating the scaling factor and performing the feature scaling. For calculating the scaling factor, we recursively construct a decision tree. As shown in Figure \ref{fig:scaling-factor}, the features lying at the top of the tree get a high scaling factor, and features at the bottom get low scaling factor. In the second phase, the features are transformed with a modified robust scaler where the scaling factor calculated in the first step, gets multiplied. Let the original dataset be represented with $D$ and the scaled dataset with $D\_transform$. As presented in Algorithm \ref{alg:algo}, the DTization takes dataset $D$ as input and returns the transformed dataset $D\_transform$ that is scaled using supervised manner.

\begin{algorithm}
\caption{Feature Scaling with DTization} 
\label{alg:algo}
\begin{algorithmic}[1]
\STATE \textbf{Input :}
 $D$ = $\{x_1,x_2,x_3....x_N\}$
\STATE \textbf{Input :}
 $D\_transform$ = $\{x_1,x_2,x_3....x_N\}$
 \STATE \textbf{Method :} DTization(D):
 \STATE $nf$= Number of features in D
 \STATE $x$ = $ \frac{log(2)}{nf} $
 \STATE $S$=$\emptyset$
 \STATE CalculateSF($D$,$x$,$S$,1)
 \STATE $D\_transform$ =$D$
 \FOR{each feature $f$ $\in$ $D$}
 \STATE $q1$= first quartile of $D[f]$
 \STATE $q3$= third quartile of $D[f]$
 \STATE $D\_transform[f]$= $S[f]$ $\times$ ($D[f]$-$q1$) $\times$ ($D[f]$-$q3$)
 \ENDFOR
 \RETURN $D\_transform$
\end{algorithmic}
\end{algorithm}
The first task of DTization feature scaling method is to calculate the feature scaling factor. The scaling factors lie in the range of 0 to 1. According to the experiment, an exponential decrease in the scaling factor as the level of the decision tree increases produces better results. Let the number of features in dataset $D$ be represented with $nf$. We employ an exponent, $x$, that makes sure the value of the equation does not exceed 1. Therefore the equation becomes $e^{x \times nf}$. Since the value of $nf$ $\geq$ 1 and $x$ $\geq$ 0, any value with the constraint will produce output greater than 1. However, we want the height value of the equation to be 1. Therefore, we deduct 1 from the equation to meet the constraint. After solving the Equation, as presented in Equation \ref{eq:sf}, we obtain the value of the exponent $x$.

\begin{equation}
\label{eq:sf}
\begin{split}
        e^{x \times nf} - 1 = 1 \\
e^{x \times nf}=2 \\
log(e^{x \times nf})=log(2) \\
x \times nf = log(2) \\
x= \frac{log(2)}{nf}
\end{split}
\end{equation}

A container $S$ is used for storing the value of the scaling factors. This process takes place in Algorithm \ref{alg:sf-calculate}. In this process, we first select the best attribute using the gini index. Gini index is used for finding out the best attribute in a dataset in C4.5 decision tree induction algorithm \cite{javed2022performance}. Let the best attribute in a sub-dataset be denoted with $f$ and the current level of DT with $d'$. Therefore, the scaling factor of feature $f$ is e$^{x \times d'}$. The dataset is then divided into two sub-datasets based on the splitting point and recursively calculated scaling factors on those datasets. The recursion terminates when the leaf node is reached.

After calculating the scaling factor, the process again comes back to Algorithm \ref{alg:algo}. Then we create a fresh dataset, $D\_transform$ by copying the original dataset. This step can be eliminated if we want the original dataset to be transformed. After that, for every feature, we calculate the first and third quartile which is used for robust scaling. Subsequently, the scaling factor is multiplied with the robust scaler output to get the desired result that is more precise and scaled depending on the feature importance. Finally, the transformed dataset is returned as output.

\begin{algorithm}
\caption{Calculate Scaling Factor} 
\label{alg:sf-calculate}
\begin{algorithmic}[1]
\STATE \textbf{Input :}
 D = $\{x_1,x_2,x_3....x_N\}$,  x = exponent, S = Scaling Factor Container
 d`= Current Level
\STATE \textbf{Output :}
$\emptyset$
\STATE \textbf{Method :} CalculateSF(D, x, S, d`):
\STATE L= Number of features in D
\IF{L+1=d`}
\RETURN $\emptyset$
\ENDIF
\STATE f=Feature with the highest gini index in D
\STATE split\_point= best splitting condition on D[f]
\STATE S[f]=$e^{x\times d`}$
\STATE D`=D[f $<$ split\_point]-D[f]
\STATE CalculateSF(D`,x,S,d`+1)
\STATE D`=D[f $\geq$ split\_point]-D[f]
\STATE CalculateSF(D`,x,S,d`+1)
\RETURN $\emptyset$
\end{algorithmic}
\end{algorithm}

\subsection{Asymptotic Analysis}
Let n denote the number of instances in a dataset and d denote the number of dimension. The first part of DTization, the scaling factor calculation, consumes $\mathcal{O} (nd)$ time complexity since it needs to construct a decison tree. Finding the value of gini index takes $\mathcal{O} (n)$ time complexity and since there is no pruning applied, it repeats the operation d times. Therefore the time complexity becomes $\mathcal{O} (nd)$. The space complexity is the same as decision tree that is $\mathcal{O} (d)$. 
\begin{equation}
\label{eq:complexity}
    \begin{split}
    \text{scaling factor calculation}= \mathcal{O} (nd) \\
    \text{feature scaling}= \mathcal{O} (nd) \\
    \mathrm{total} = \mathcal{O} (nd+nd) \\
    \mathrm{total} = \mathcal{O} (2\times(nd)) \\
    \mathrm{total} = \mathcal{O} (nd) \\
    \end{split}
\end{equation}

While transforming the dataset, it needs to iterate all the values of the dataset that takes $\mathcal{O} (nd)$ time. Regarding the space complexity, it is $\mathcal{O} ((nd))$ since it needs to hold the transformed dataset.
\begin{equation}
\label{eq:space}
    \begin{split}
    \text{scaling factor calculation}= \mathcal{O} (d) \\
    \text{feature scaling}= \mathcal{O} (nd) \\
    \mathrm{total} = \mathcal{O} (d+nd) \\
    \text{ $d$ $\ll$ $nd$} \\ 
    \mathrm{total} = \mathcal{O} (nd) \\
    \end{split}
\end{equation}

Therefore, as shown in Equation \ref{eq:complexity} and Equation \ref{eq:space}, the overall time complexity is $\mathcal{O} (nd)$ and space complexity is also $\mathcal{O} (nd)$ which is equal to the other feature scaling techniques.

\section{Results}
\label{sec:results}
This section discusses the outcomes obtained from this algorithm along with a comparison with state-of-the-art feature scaling techniques.

\subsection{Experimental Setup}
We conducted the experiment on Kaggle. The experiment utilized the Python programming language along with three Python libraries named Pandas, scikit-learn, and Numpy. 

\subsection{Evaluation Matrices}
For evacuating the classification datasets, five evaluation matrices were selected. These are the accuracy, precision, recall, F1-score, and Matthew's correlation coefficient (MCC). Accuracy measures the percentage of correctly predicted outcomes. Precision accounts for the true positive predictions out of the positive predictions. Recall, however, measures the true positive predictions out of actual positive predictions. F1-score measures the mean of both precision and recall. Finally, the most reliable MCC measures the correctness of the model in all four coordinates in the confusion matrix \cite{chicco2023matthews}. Therefore, it is considered the most reliable matrix among the five metrics compared \cite{islam2023novel}.

\begin{equation}
\text { Accuracy }=\frac{\mathrm{TP}+  \mathrm{TN}}{\mathrm{TP}+\mathrm{PP}+\mathrm{TN}+\mathrm{FN}}
\end{equation}

\begin{equation}
\text { Precision }=\frac{\mathrm{TP}}{\mathrm{TP}+\mathrm{FP}}
\end{equation} 

\begin{equation}
\text { Recall }=\frac{\mathrm{TP}}{\mathrm{TP}+\mathrm{FN}}
\end{equation}

\begin{equation}
 \text { F1-score }=\frac{2 \times \text { Precision } \times \text { Recall }}{\text { Precision }+ \text { Recall }}
\end{equation}

\begin{equation}
\mathrm{MCC}=\frac{\mathrm{TP} \times \mathrm{TN}-\mathrm{FP} \times \mathrm{FN}}{\sqrt{(\mathrm{TP}+\mathrm{FP})(\mathrm{TP}+\mathrm{FN})(\mathrm{TN}+\mathrm{FP})(\mathrm{TN}+\mathrm{FN})}}
\end{equation}

For the regression datasets, three matrices were used that is the mean absolute error (MAE), mean squared error (MSE) and R-squared (or R$^{2}$) error. MAE calculates the average absolute difference between predicted and actual values that describes the average magnitude of errors. MSE calculates the average of squared differences between predicted and actual values leading to a more reliable measurement for noisy data. The R-squared error determines the proportion of the dependent variable's variance that the independent variable can account for \cite{shen2022wind}.

\begin{equation}
\label{eq:mae}
    \mathrm{MAE} = \frac{1}{n} \sum_{i=1}^{n} |y_i - \hat{y}_i|
\end{equation}
\begin{equation}
    \mathrm{MSE} = \frac{1}{n} \sum_{i=1}^{n} (y_i - \hat{y}_i)^2
\end{equation}

\begin{equation}
    \mathrm{R^2} = 1 - \frac{{\sum_{i=1}^{n} (y_i - \hat{y}_i)^2}}{{\sum_{i=1}^{n} (y_i - \bar{y})^2}} 
\end{equation}

\subsection{Datraset Selection And Preprocessing}

To compare the effectiveness of the proposed algorithms, five classification datasets, prsented in Table \ref{tab:classification-datasets}, were carefully selected for this study. The datasets chosen were whine, yeast, fetal\_health, sepsis, and Credit Card. Each dataset is a representation of a distinct problem domain, offering various difficulties for classification tasks. The whine dataset involves predicting one of three whine categories and consists of 178 samples with 13 features. The yeast is a biological dataset consists of 1484 samples and eight features that are intended to be categorized into ten different classes. The 2126 samples in the fetal\_health dataset have a total of 21 features, and it also involves predicting one of three classes. The sepsis dataset consists of medical data of 110,204 subjects admitted in Norway from 2011 to 2012. The credit card dataset, which includes 284,807 samples and 30 features, has a large amount of data and is used to predict whether a transaction is fraudulent or not, creating a binary classification problem. Notably, the imbalanced class distributions in the sepsis and credit Card datasets pose extra difficulties for classification algorithms. 
\begin{table}[!ht]
    \centering
    \caption{Classification Datasets Used for Evaluation}
    \begin{tabular}{|p{2.5cm}|p{1.5cm}|p{1.5cm}|l|}
    \hline
        \textbf{Dataset Name} & \textbf{Samples} & \textbf{Features} & \textbf{\# of classes} \\ \hline
        Whine & 178 & 13 & 3 \\ \hline
        Yeast & 1484 & 8 & 10 \\ \hline
        Fetal\_health & 2126 & 21 & 3 \\ \hline
        Sepsis & 110204 & 4 & 2 \\ \hline
        Credit Card & 284807 & 30 & 2 \\ \hline
    \end{tabular}
    \label{tab:classification-datasets}
\end{table}

Several datasets were chosen for the regression analysis in order to compare the effectiveness of various machine learning algorithms. NBA player statistics, possum, medical costs, house price prediction, and productivity were the selected datasets. The possum dataset had 101 samples and 11 features, whereas the NBA player stats dataset had 539 samples and 26 features. The house price prediction dataset had 21,613 samples with 18 features, and the medical Cost dataset had 1338 samples with 3 features. Lastly, the productivity dataset contained 691 samples and 10 features. Three non-numeric columns from the NBA dataset, three from the possum dataset, three from the medical cost dataset, two columns from the house price prediction dataset, and four columns from the productivity dataset were removed to ensure compatibility with the algorithms used. The datasets are further described in Table \ref{tab:regression-datasets}.

\begin{table}[!ht]
    \centering
    \caption{Regression Datasets Used for Evaluation}
    \begin{tabular}{|l|l|l|p{2cm}|}
    \hline
        \textbf{Dataset Name} & \textbf{Samples} & \textbf{Features} & \textbf{Target Variable Range} \\ \hline
        NBA player stats & 539 & 29 & 1282 \\ \hline
        possum & 101 & 14 & 68 \\ \hline
        Medical Cost & 1338 & 6 & 61470.99919 \\ \hline
        House Price Prediction & 21613 & 20 & 7625000 \\ \hline
        Productivity & 691 & 14 & 0.866778442 \\ \hline
    \end{tabular}
    \label{tab:regression-datasets}
\end{table}

\subsection{Comparison with Other Feature Scaling Techniques}
To compare the result of the proposed method, we selected four popular feature scaling methods named min-max scaler, standard scaler, log transformation, and robust scaler. \par 
The min-max scaler is used to transform the input features into a specified range. In our experiment, we transformed the features into the range of 0 to 1. The equation of min-max scaler is presented in equation \ref{eq:minmax} where the input variable is represented with $X$, the minimum value of the input variable with $X_{min}$ and the maximum value with $X_{max}$ and finally the scaled output with $X_{scaled}$.
\begin{equation}
\label{eq:minmax}
X_{\text{scaled}} = \frac{X - X_{\text{min}}}{X_{\text{max}} - X_{\text{min}}}
\end{equation}

Standard scaler transforms the data to have a mean ($\mu$) of 0 and a standard deviation ($\sigma$) of 1. The equation of this scaler is presented in Equation \ref{eq:sc};
\begin{equation}
\label{eq:sc}
X_{\text{scaled}} = \frac{X - \mu}{\sigma}
\end{equation}

The log transformation is a very straightforward method, as shown in equation \ref{eq:log}, that passes the input through a log function.
\begin{equation}
\label{eq:log}
X_{\text{scaled}} = \log(X)
\end{equation}

Robust scaler scales the data by subtracting the value of the first quartile and dividing it by the difference between first and third quartile ranges. The equation of robust scaler is pacified in Equation \ref{eq:robust}.
\begin{equation}
\label{eq:robust}
X_{\text{scaled}} = \frac{X - \text{Q1}(X)}{\text{Q3}(X)- \text{Q1}(X)}
\end{equation}

\subsection{Result Analysis and Discussion}

The performance of DTization was evaluated on the selected classification datasets. The K-Nearest Neighbors (KNN) algorithm was selected for evaluating the classification datasets since it is vulnerable to feature scaling. The value of nearest neighbors, k, was selected a small value, 3, to make the algorithm more sensitive to scaling. In sepsis, the imbalance dataset, KNN achieved an MCC of 0.0631 using the other feature scaling approaches mentioned above. However, with the DTization approach, it achieved improved results with an MCC of 0.0718. In credit card, another imbalance dataset, the DTization improved the MCC from 0.28 to 0.83. On whine, the dataset with the least number of samples, the proposed feature scaling improved the accuracy from 75\% to 94\%. In the fatal\_health dataset, the proposed algorithm scored the same results as the other techniques. The noteworthy finding from the performance evaluation is that none of the other feature scaling matrices tested resulted any improvements in the performance of the selected datasets. The detailed results of the classification datasets are present in Table \ref{tab:classification}.

\begin{table}[h]
    \centering
    \caption{Performance Comparison on Classification Datasets}
    \begin{tabular}{|p{1.2cm}|p{1cm}|p{0.93cm}|p{0.9cm}|p{0.66cm}|p{0.7cm}|p{0.6cm}|}
    \hline
        \textbf{Dataset Name} & \textbf{Approach} & \textbf{Accuracy} & \textbf{Precision} & \textbf{Recall} & \textbf{F1-score} & \textbf{MCC} \\ \hline
        Whine & Others & 0.7500 & 0.7188 & 0.7177 & 0.7174 & 0.6173 \\ 
        ~ & DTization & 0.9444 & 0.9521 & 0.9407 & 0.9448 & 0.9157 \\ \hline
        Yeast & Others & 0.5522 & 0.4889 & 0.4897 & 0.4774 & 0.4206 \\ 
        ~ & DTization & 0.5859 & 0.4151 & 0.3984 & 0.4014 & 0.4637 \\ \hline
        Fetal\_health & Others & 0.8967 & 0.8055 & 0.8244 & 0.8120 & 0.7135 \\ 
        ~ & DTization & 0.8967 & 0.8055 & 0.8244 & 0.8120 & 0.7135 \\ \hline
        Sepsis & Others & 0.9112 & 0.5189 & 0.5527 & 0.5213 & 0.0631 \\ 
        ~ & DTization & 0.9131 & 0.5208 & 0.5619 & 0.5240 & 0.0718 \\ \hline
        Credit Card & Others & 0.9982 & 0.5454 & 0.9537 & 0.5822 & 0.2872 \\ 
        ~ & DTization & 0.9994 & 0.8727 & 0.9657 & 0.9140 & 0.8332 \\ \hline
    \end{tabular}
    \label{tab:classification}
\end{table}

The regression datasets are evaluated with linear regression algorithm. Table \ref{tab:regression} presents the performance comparison on the regression datasets. Similar to the classification datasets, other feature scaling techniques made no changes in the overall performance of any of the datasets. Although the proposed method resulted in a degradation of MAE in two datasets (Possum and Medical Cost), a significant improvement in the MAE was observed in the other datasets. Particularly on the Productivity dataset, DTization resulted in a 63\% improvement in the performance. Overall, the proposed method was found to be a fruitful approach for ameliorating the performance of a regression models.

\begin{table}[h]
    \centering
    \caption{Performance Comparison on Regression Datasets}
    \begin{tabular}{|p{2cm}|l|p{1.1cm}|p{1.88cm}|p{0.8cm}|}
    \hline
        \textbf{Dataset Name} & \textbf{Approach} & \textbf{MAE} & \textbf{MSE} & \textbf{R-squared} \\ \hline
        NBA player stats & Others & 97.719 & 19660.199 & 1.311 \\ 
        ~ & DTization & 90.269 & 16038.065 & 0.947 \\ \hline
        Possum & Others & 2.008 & 7.012 & 1.628 \\ 
        ~ & DTization & 2.421 & 7.782 & 1.637 \\ \hline
        Medical Cost & Others & 9677.197 & 167944752.131 & 1.823 \\ 
        ~ & DTization & 10035.112 & 189371619.776 & 2.682 \\ \hline
        House Price Prediction & Others & 163956.522 & 66988766731.256 & 0.126 \\ 
        ~ & DTization & 98406.645 & 32732368119.092 & 0.676 \\ \hline
        Productivity & Others & 0.090 & 0.018 & 0.564 \\ 
        ~ & DTization & 0.033 & 0.004 & 0.807 \\ \hline
    \end{tabular}
    \label{tab:regression}
\end{table}

\subsection{Comparison With Existing Works}
A comparison with related works discussed in Section \ref{sec:rw} is presented in Table \ref{tab:compare}. According to the comparison, none of the recent research articles address the need for a universal supervised feature scaling method. Therefore, we have proposed a supervised feature scaling method that can be applied for both classification and regression tasks to achieve a higher performance. 

\begin{table*}
    \centering
    \caption{Comparison with related works}
    \begin{tabular}{|p{2cm}|p{3cm}|p{3cm}|p{3cm}|p{3cm}|}
    \hline
        \textbf{Paper} & \textbf{Algorithms Used} & \textbf{Tasks Performed} & \textbf{Application Area} & \textbf{Feature Engineering Type} \\ \hline
        Ahsan et al.\cite{ahsan2021effect} & CART, Robust Scaler, Quantile Transformer & Classification & Heart Disease Classification & Feature Scaling \\ \hline
        Zhang and Zhao \cite{zhang2017fetal} & PCA, AdaBoost & Classification & Featal State Detection & Feature Selection \\ \hline
        Devi et al. \cite{devi2021variant} & PCA, LDA, SVM, Naive Bayes, Logistic Regression & Classification & Hypothyroid Disease Classification & Feature Selection \\ \hline
        Md et al. \cite{md2023enhanced} & Scaling, Feature Selection & Classification & Liver Disease Classification & Feature Scaling, Feature Selection \\ \hline
        Tripathi et al. \cite{tripathi2021feature} & Robust Scaler, Chi-square, Random Forest & Classification & Malware Detection & Feature Scaling, Feature Selection \\ \hline
        Shekhar et al. \cite{shekhar2022pknn} & Mutual Information, KNN & Classification, Regression & All & Feature Selection \\ \hline
        Albulayhi et al. \cite{albulayhi2022iot} & Information Gain, Gain Ration, Set Theory & Classification & Intrusion Detection & Feature Selection \\ \hline
        Proposed & Decision Tree, Robust Scaler & Classification, Regression & All & Feature Scaling \\ \hline
    \end{tabular}
    \label{tab:compare}
\end{table*}

\section{Conclusion}
\label{sec:conclusion}
In this paper, we have presented a new approach of feature scaling. The proposed solution incorporates decision tree algorithm to find out the feature importance. Based on the importance, each future gets scaled differently with the help of robust scaler. The proposed solution consumes the same asymptotic time and space complexity, however, produces better results compared to the state-of-the-art feature scaling methods. The experimental results show that the model can be a considerable option for enhancing the machine learning model's performance. 


\section*{Supplementary Materials}
The Python codes for the scaller along with the datasets are present in this link: \url{https://github.com/NifulIslam/DTization}. The README.md file presents the detailed manual for using the scaler. 

\bibliographystyle{IEEEtran}
\bibliography{FILES/bibfile.bib}

\end{document}